# Detecção de comunidades em redes complexas para identificar gargalos e desperdício de recursos em sistemas de ônibus

**Carlos Caminha, Vasco Furtado, Vládia Pinheiro, Caio Ponte**

Laboratório de Engenharia do Conhecimento – Universidade de Fortaleza (UNIFOR) – Caixa Postal 60811-905 – Edson Queiroz, Fortaleza – CE – Brazil

***Abstract.*** *We propose here a methodology to help to understand the shortcomings of public transportation in a city via the mining of complex networks representing the supply and demand of public transport. A process of characterization of the supply and demand networks of the bus system of a large Brazilian metropolis was conducted and it also shed light on the potential overload of demand and waste in the supply of resources that can be mitigated through strategies of supply and demand balance.*

***Resumo.*** *Aqui se propõe uma metodologia para ajudar a compreender as deficiências do transporte público através da mineração de redes complexas que representam a oferta e a demanda de transportes públicos. Foi conduzido um processo de caracterização de redes de oferta e demanda do sistema de ônibus de uma grande metrópole brasileira e o mesmo lançou uma luz sobre potencial sobrecarga da demanda e desperdício na oferta de recursos que podem ser mitigados com estratégias de equilíbrio entre oferta e demanda.*

## 1. Introdução

A era da informação digital trouxe a necessidade de, a partir de dados diversos e volumosos, criar meios para gerar conhecimento e aplicá-lo com eficácia. Uma das áreas que melhor exemplificam esse contexto é a mobilidade urbana. Sensores e meios digitais registram informações diárias sobre trilhas de pessoas, tornando-se uma entrada rica para a realização de estudos para a compreensão da mobilidade humana e seu impacto no transporte público.

Um exemplo de sistema largamente sensoriado é o sistema de ônibus de uma grande metrópole. Com sensores como GPS (*Global Positioning System*), que registram por onde os veículos trafegam, e sistema de bilhetagem, que registram o pagamento e tarifas por parte dos passageiros, são gerados diariamente imensos bancos de dados e um grande desafio é extrair informações que tenham potencial de identificar falhas que deixam usuários do sistema insatisfeitos, como por exemplo a superlotação de veículos.

Ao longo dos anos inúmeros trabalhos trataram de analisar oferta e demanda de sistemas de ônibus [Chang and Schonfeld 1991], [Oppenheim 1995], [Domencich and McFadden 1975] e dois aspectos se mostraram particularmente desafiadores ao avaliar ocupação extrema (veículos lotados ou vazios). Primeiro, a ausência de sensores que permitam que a demanda seja mensurada por completo. Em muitos sistemas de bilhetagem só é registrado o momento em que o usuário sobe nos veículos, sendo necessário fazer uso de heurísticas para estimar onde o usuário desceu [Gordillo 2006],

[Hua-ling 2007]. Essa limitação dificulta a avaliação da lotação dos veículos, obrigando analistas a comparar a oferta, que é completamente conhecida, com a demanda que muitas vezes é analisada com amostras pouco significativas. O segundo desafio ao avaliar ocupação extrema em redes de transporte coletivo é que a complexidade das conexões desses sistemas dificulta a percepção de gargalos. Nesse tipo de sistema é comum avaliar conexões fracas de oferta como possíveis pontos de gargalo, no entanto somente uma avaliação topológica dos subcomponentes que cercam essas arestas pode confirmar a existência do problema [GAO et al 2005].

Diante dessa problemática, uma estratégia que se mostra adequada para compreensão desse tipo de fenômeno é a utilização de redes complexas. O potencial desse tipo de instrumento para tratar problemas que exigem a necessidade de abstração de aspectos como o desconhecimento de parte dos dados, aliado a riqueza de métricas e algoritmos já existentes no estado da arte, justificam sua aplicação para resolver uma série de problemas. Observa-se recentemente uma quantidade significativa de trabalhos que modelam redes para compreensão de sistemas complexos no contexto de transporte aéreo [Wang et al. 2011], ferroviário [Lenormand et al. 2014], urbano [Munizaga and Palma 2012] e bicicletas [Hamon et al. 2013]. Em geral, estes estudos visam caracterizar redes utilizando métricas como a distribuição de peso, comprimento médio do caminho e coeficiente de agrupamento. O trabalho de [Sienkiewicz and Hołyst 2005], por exemplo, comparou o sistema de transporte público em 22 cidades da Polônia. Entre outras características, os autores mostraram que a distribuição de graus e pesos segue uma lei de potência hierarquicamente organizada.

O trabalho apresentado em [Elmasry and McCann 2003] é particularmente relevante neste contexto e é complementar ao nosso. O artigo apresenta um algoritmo para detectar gargalos em uma rede congestionada de grande escala comutada por pacotes. A abordagem baseia-se na estimativa do valor esperado de atraso em trechos da rede (trechos estes representados por arestas de um grafo) e na conversão deste valor esperado para um peso por ligação.

Neste artigo, descreveremos nossa experiência em Fortaleza, uma grande metrópole brasileira, no âmbito do seu projeto para *Smart City*. Usando uma rede de oferta reconstruída a partir de dados de GPS de ônibus em [Caminha et al. 2016], e uma rede que representa a demanda do sistema de ônibus na cidade estimada a partir de um algoritmo heurístico em [Caminha et al. 2017], foi realizada uma caracterização que revelou um desequilíbrio global do sistema. Além disso, uma estratégia baseada em detecção de comunidades nos levou a identificar trechos da rede de transporte público com problemas, como gargalos e/ou desperdício de recursos.

## 2. Conjunto de dados

A rede de oferta pode ser definida como um grafo dirigido, *G (V, E)*, com vértices, *v (∈ V)*, representando as paradas e arestas de ônibus, *e (∈ E)*, entre dois pontos de ônibus. Formalmente o peso da aresta, $w_{v_i \rightarrow v_j}$, representa a oferta de ônibus entre duas paradas de ônibus $v_i$ e $v_j$ *(∈ V)*. Esta oferta foi calculada a partir do somatório do peso das linhas de ônibus, $w_{Li}$, que passam por dois pontos de ônibus. Formalmente $w_{v_i \rightarrow v_j} = \sum_{i=1}^{N} w_{Li}$, onde *N* é o número total de linhas de ônibus que visitam, em sequência, $v_i$ e $v_j$. Por sua vez, o peso das linhas de ônibus, é calculado a partir do produto da quantidade de veículos alocados na linha *Li*, $V_{Li}$, pelo número de viagens que cada veículo faz em um dia em *Li*,

$C_{Li}$. Ou seja, $w_{Li} = V_i C_i$. Ao todo, essa rede possui 4783 vértices e 5876 arestas e foi produzida em [Caminha et. al. 2016].

A rede de demanda foi produzida em [Caminha et al. 2017]. No processo de reconstrução dos caminhos dos usuários foi suposto que as trajetórias dos usuários de ônibus são definidas pela composição das rotas dos ônibus que os levam entre seu pares origem-destino. Neste contexto, foram reconstruídas as trajetórias de usuários de ônibus como um grafo dirigido $G\ (V, E)$, onde $V$ e $E$ são o conjunto de vértices $v$ e arestas $e$, respectivamente. Uma aresta e entre os vértices $v_i$ e $v_j$ é definida pelo par ordenado ($v_i$, $v_j$). Nessa rede, os vértices representam paradas de ônibus e as arestas representam a demanda de usuários de ônibus entre dois pontos de ônibus consecutivos. Para cada aresta, $e$, que liga um par ordenado ($v_i$, $v_i$) é definido um peso, $w_{Eij}$, que soma o total de usuários que passaram por $e$. Ao todo, essa rede possui os mesmos 4783 vértices e 5876 arestas existentes na rede de oferta. O único aspecto que difere as redes é o peso das arestas, na rede de oferta esse peso é mensurado em quantidade de veículos que ofertam rotas e na rede de demanda o mesmo é mensurada em quantidade de pessoas que passaram dentro dos ônibus. Vale ainda ressaltar que a rede de demanda representa apenas uma amostra da necessidade da população, mesmo assim [Caminha et al. 2017] argumentam que sua amostra representa 40% da demanda de ônibus da cidade, uma das maiores amostras de demanda de sistema de ônibus que se tem notícia.

Tanto a rede de oferta quanto a rede de demanda possuem dados do dia 11 de março de 2015. Uma quarta-feira, dia útil na cidade. A Figura 1 ilustra a disposição geográfica das paradas de ônibus e suas conexões.

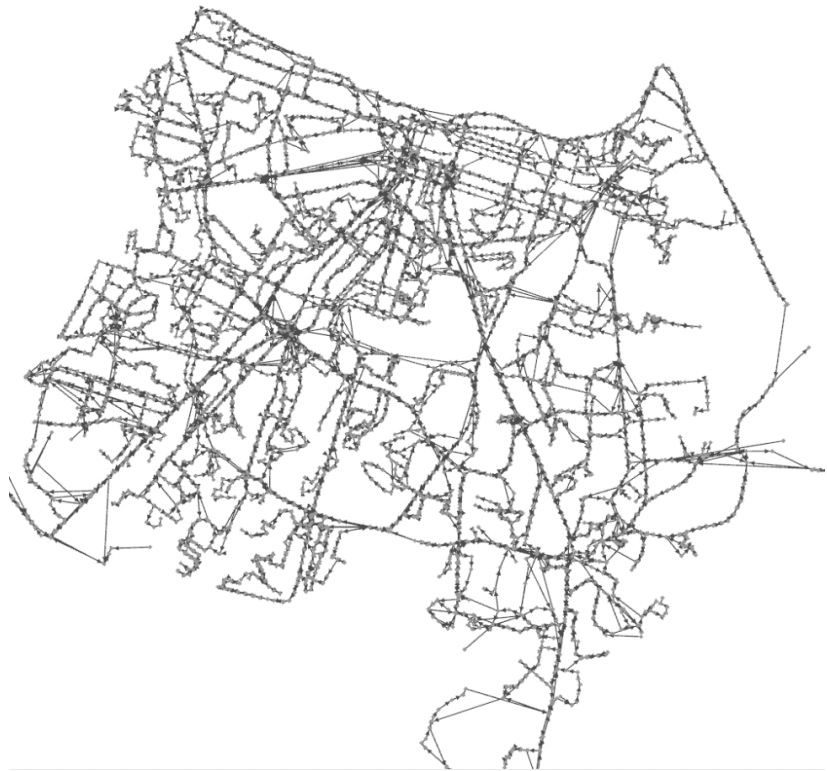

Figura 1. Estrutura da rede de ônibus de fortaleza. Os vértices foram georreferenciados para melhor visualização.

# 3. Caracterização de redes de oferta e demanda para encontrar desequilíbrios globais

Nesta seção será descrito o processo de caracterização das redes oferta e demanda do sistema de ônibus de Fortaleza. Para permitir comparações, os pesos das arestas das duas redes foram normalizados. Na rede de demanda, para cada aresta, $e_d$, o peso, $w_d$, que descreve a demanda de passageiros entre dois pontos de ônibus, foi normalizado pela razão $w_d/w_{dmax}$ onde $w_{dmax}$ é o maior peso registrado na rede. Para a rede de oferta utilizou-se a mesma estratégia para calcular o peso que, nesse caso, representa a quantidade de ônibus disponíveis. Note-se que ambas as redes têm a mesma distribuição de graus, isto é, para cada aresta na rede de oferta há uma aresta correspondente na rede de demanda, essas redes diferem somente em relação aos seus pesos de arestas.

A Figura 2 ilustra a distribuição de pesos das arestas para as duas redes. E (a) observa-se uma lei de potência [Barabási et al. 2000] [Clauset et al. 2009] com exponente α = -2.90 para a rede de oferta. Este resultado sugere que esta rede foi projetada para oferecer um elevado número de recursos em poucos lugares e poucos recursos em vários locais. Da mesma forma, uma lei de potência pode ser observada para a rede de demanda na Figura 2 (b). O exponente α = -1.97 indica que há poucos lugares com alta demanda e vários lugares com baixa demanda.

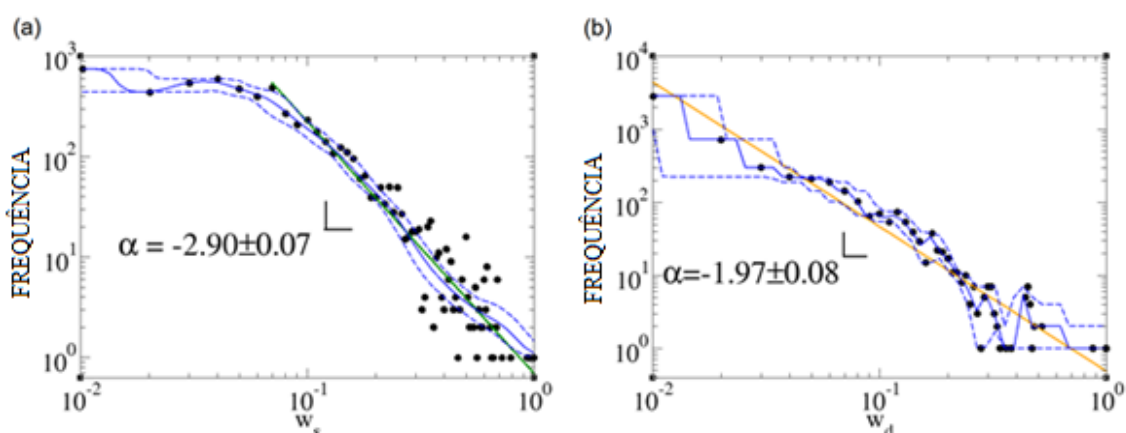

Figura 2. Distribuição de pesos das arestas das redes de oferta e demanda. Em (a), a linha verde mostra a regressão aplicada aos dados da rede de oferta. Em (b) a regressão é representada pela linha laranja para os dados da demanda. As linhas tracejadas azuis representam os intervalos de confiança estimados com o método de Nadaraya Watson [Nadaraya 1964] [Watson 1964]. As distribuições foram geradas usando 50 *beans* logarítmicos.

Apesar das distribuições apresentarem similaridade em termos de escala, isso não implica necessariamente que as redes estejam equilibradas. A primeira sugestão de desequilíbrio é a diferença entre os expoentes, o volume de concentração em poucos lugares é muito mais intenso na rede de oferta. A Figura 3 ilustra a diferença na inclinação das retas estimadas.

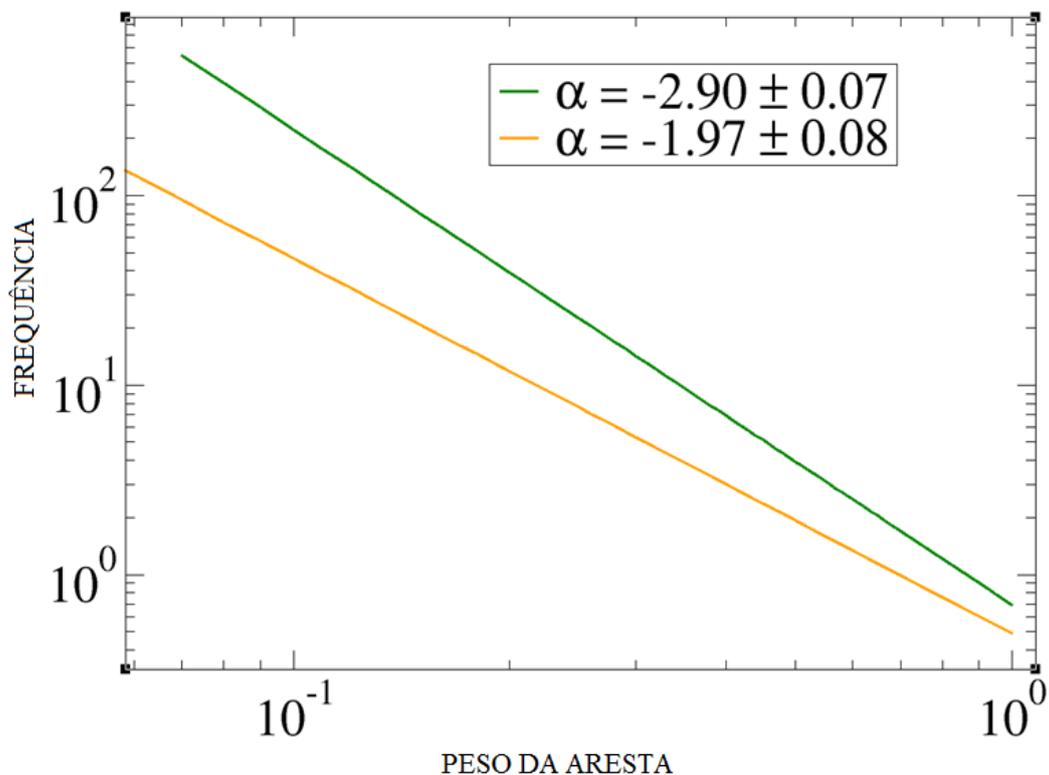

Figura 3. Comparação entre as escalas das leis de potência encontradas. A linha verde ilustra a regressão estimada na rede de oferta e a linha laranja é a regressão estimada para a distribuição dos pesos das arestas da rede de demanda.

Ainda na distribuição de pesos das arestas das redes, na Figura 4 é ilustrada a distribuição cumulativa desses dados. A curva laranja mostra que a rede de demanda acumula mais arestas com pesos pequenos que a rede de oferta, as probabilidades são iguais nas duas redes e arestas com peso 0,16. Essa é outra sugestão de desequilíbrio devido às diferenças entre as probabilidades acumuladas em ambas as redes.

Foi investigado também se as redes poderiam ser divididas em comunidades com alto coeficiente de agrupamento e baixa conectividade com outras comunidades. Para a rede de oferta isso pode indicar a existência de áreas onde o suprimento de ônibus é privilegiado (dentro das comunidades) e também gargalos (arestas fracas) entre essas áreas. Foi utilizado o algoritmo [Blondel et al. 2008] para detectar as comunidades. Esse algoritmo faz uso de um método heurístico baseado na otimização da modularidade de grafos. A modularidade de um conjunto de nós é medida por um valor real entre 0 e 1, sendo calculada pela relação entre a quantidade de arestas que conectam os elementos do conjunto entre si pelo total das arestas do conjunto de nós [Girvan 2002] [Wilkinson 2004]. O algoritmo pode ser dividido em duas fases que são repetidas iterativamente. Na primeira fase é atribuída uma comunidade diferente para cada nó de rede. Assim, nesta divisão inicial há tantas comunidades quanto nós. Para cada nó $i$ é considerado cada um de seus vizinhos $j$, e é avaliado o ganho de modularidade ao mover $i$ de sua comunidade para a comunidade de $j$. O nó $i$ é então colocado na comunidade para o qual esse ganho é máximo, mas somente se esse ganho for positivo. Se nenhum ganho positivo é possível, o nó $i$ permanece em sua comunidade original. Este processo é aplicado a todos os nós repetidamente até que nenhuma melhoria possa ser alcançada e assim a primeira fase é completada. A segunda fase do algoritmo consiste na construção de uma nova rede cujos nós são comunidades encontradas durante a primeira fase. Para fazer isso, os pesos das conexões entre os novos nós são dados pela soma do peso das conexões entre os nós nas duas comunidades correspondentes. As conexões entre os nós da mesma comunidade são representadas por auto relacionamentos. As fases são repetidas até que a modularidade global máxima seja encontrada.

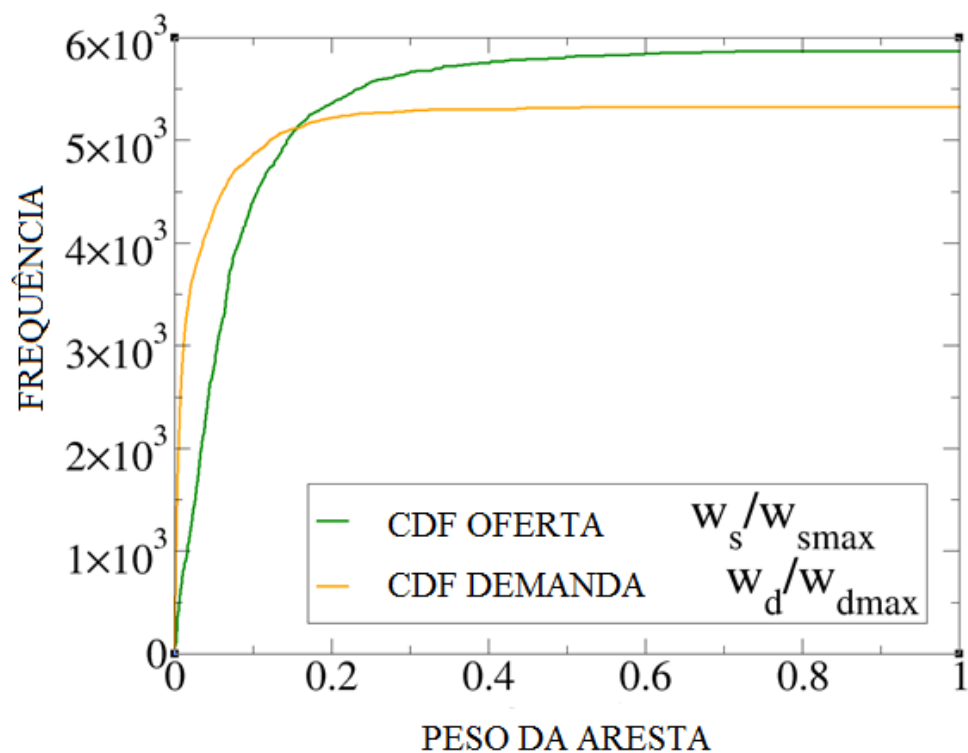

Figura 4. Distribuições cumulativas dos pesos das arestas nas redes estudadas. A curva laranja representa a distribuição cumulativa dos pesos da demanda e a curva verde representa a distribuição cumulativa correspondente a rede de oferta.

A Figura 5 ilustra o aumento da modularidade quando a rede é dividida em mais comunidades. A modularidade da rede de oferta é maior do que a rede de demanda em praticamente todas as configurações, mais precisamente, as modularidades das redes só convergem quando o número de comunidades e inferior a 10, aproximadamente. Esse resultado revela que a rede de oferta parece ser projetada para atender a movimentos mais curtos do que o fato de que as pessoas precisam. Os valores de modularidade encontrados na rede de oferta indicam que as arestas dentro das comunidades dessa rede são mais fortes do que as arestas encontradas nas comunidades da rede de demanda, possivelmente porque os usuários do sistema de ônibus viajam por trechos mais logos que o esperado, aumentando o peso das arestas nesses locais e impedindo que algoritmos de detecção de comunidade encontrem configurações modulares como as encontradas na rede de oferta.

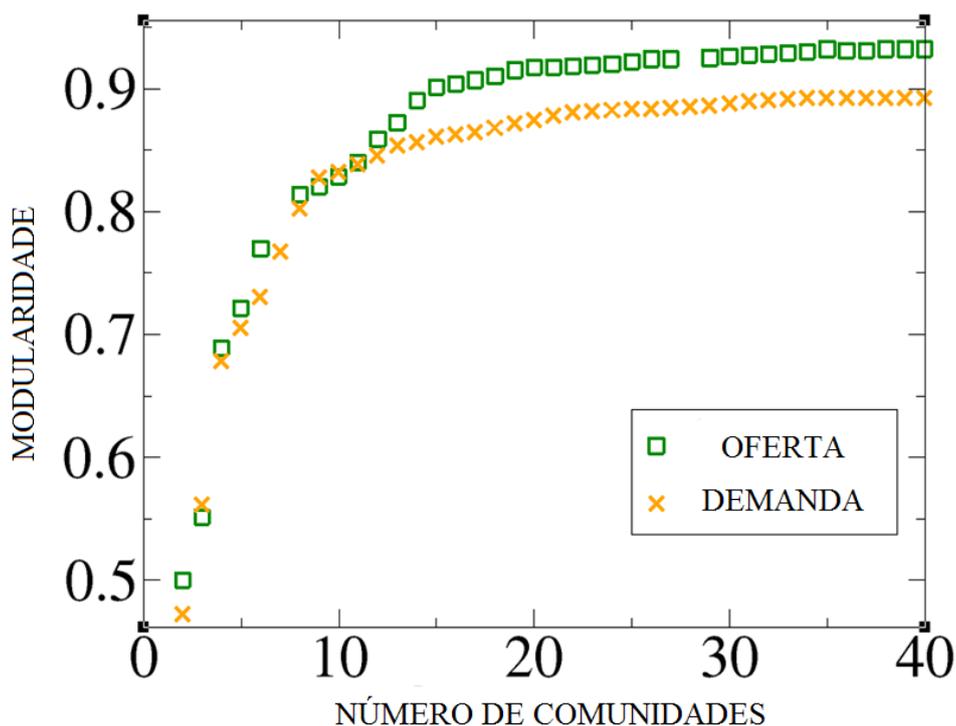

Figura 5. Comparação entre a modularidade das duas redes. A linha verde ilustra a modularidade da rede de oferta na medida em que a rede é dividida em mais comunidades. A linha laranja ilustra o comportamento de modularidade da rede de demanda.

## 5. Detecção de comunidades para encontrar gargalos e desperdício de recursos

Verificou-se que a demanda e a oferta de ônibus em Fortaleza é significativamente alterada em intervalos de três horas, com picos de uso das 5:00 às 8:00 e das 17:00 às 20:00. Por esse motivo foram geradas redes de oferta e demanda para os seguintes intervalos de tempo: De 2:01 às 5:00; 5:01 às 8:00; 8:01 às 11:00, 11:01 às 14:00; 14:01 às 17:00.; 17:01 às 20:00; 20:01 às 23:00; e das 23:01 as 2:00. O intervalo onde a rede é mais utilizada é de 5:01 às 8:00, por esse motivo vamos concentrar nossa análise neste período.

Em redes equilibradas, onde a oferta cresce em proporção direta ao crescimento da demanda, espera-se que haja uma relação isométrica entre volumes de oferta e demanda [Banavar et al 2002]. A Figura 6 ilustra a correlação entre os pesos das arestas das redes de oferta e demanda no intervalo das 5:01 às 8:00 da manhã. A baixa correlação, $R^2 = 0,53$, revela um crescimento não correlacionado, indicando mais uma vez desequilíbrio entre oferta e demanda. O valor de $\beta = 1.24$, para uma equação do tipo $Y = aX^\beta$, não revela qualquer relação isométrica [Kleiber 1961]. Acima da reta vermelha (regressão linear) estão as arestas (ou trechos da rede), onde a demanda é proporcionalmente maior que a oferta. No entanto, chama a atenção para a maior concentração de pontos abaixo da linha de regressão, indicando que há partes da cidade em que os recursos oferecidos são, proporcionalmente, maiores do que a demanda existente na região. Isso sinaliza que se pode melhor atender a demanda da rede sem a necessidade de aumentar o custo operacional da mesma, apenas tentando equilibrar a oferta em algumas partes da cidade.

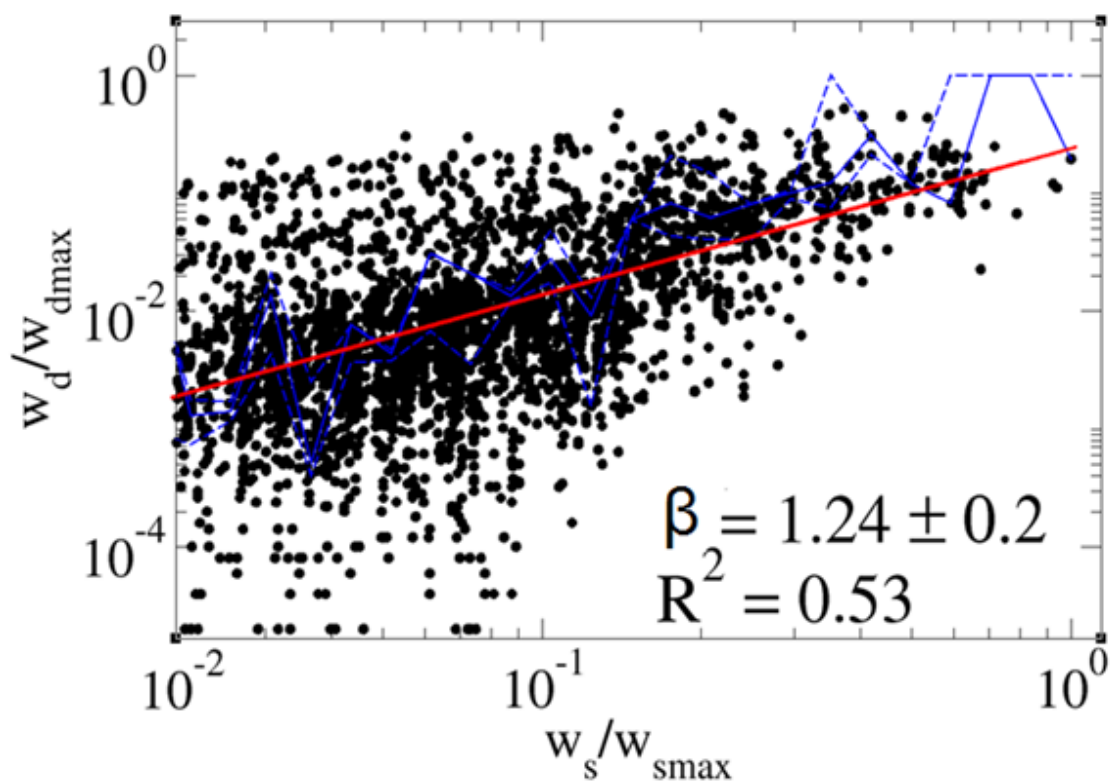

Figura 6. Correlação entre o volume de suprimento e o volume de demanda em todas as arestas da rede no intervalo das 5:00 às 8:00 da manhã. Cada ponto preto representa uma aresta. O eixo x representa o bordo de fornecimento de peso e o eixo y o peso da borda de demanda. A linha vermelha é a regressão aplicada aos dados. As linhas tracejadas a azul ilustram os intervalos de confiança estimados com o método de Nadaraya Watson. Os dados são mostrados em escala logarítmica.

Na Figura 7 (a) são ilustrados trechos da rede de oferta onde a mesma está sobrecarregada, esses trechos são percorridos por linhas de ônibus que possivelmente estão com veículos lotados. Esses trechos foram identificados a partir das arestas removidas pelo método de detecção de comunidades, detalhado na seção anterior. Uma aresta fraca removida (por um método de detecção de comunidades) da rede de oferta pode significar um gargalo de oferta, pois essa aresta conecta movimentos intercomunidades que já se mostraram predominantes [Caminha et al 2016]. Em todas as arestas removidas foi calculado o índice de sobrecarga $IS = w_s/w_{smax} - w_d/w_{dmax}$. Arestas com $IS < 0$ são trechos onde proporcionalmente a oferta e maior que a demanda, e potencialmente onde estão os gargalos. Observou-se ainda que as linhas de ônibus que possuem gargalo normalmente levam pessoas ao centro da cidade (destacado pelo círculo vermelho).

Na Figura 7 (b) são ilustrados trechos da rede onde potencialmente os ocorrem desperdício de recursos, ou seja, ônibus vazios. Esses trechos foram identificados a partir das arestas removidas pelo método de detecção de comunidades, agora na rede de demanda. Métodos de detecção de comunidades removem conexões fracas entre componentes com alto coeficiente de agrupamento, que no caso da rede de demanda são sub-regiões de fluxo intenso de passageiros. Nesse contexto, as arestas eliminadas pelo método em questão representam trechos da rede de baixo fluxo de passageiros, onde potencialmente podem ocorrer desperdícios de recursos. De forma similar ao que foi feito na oferta, calculamos o índice de desperdício $ID = w_d/w_{dmax} - w_s/w_{smax}$, onde arestas com $ID < 0$ representam trechos onde proporcionalmente a demanda e maior que a oferta, e por consequência onde estão sendo desperdiçados recursos.

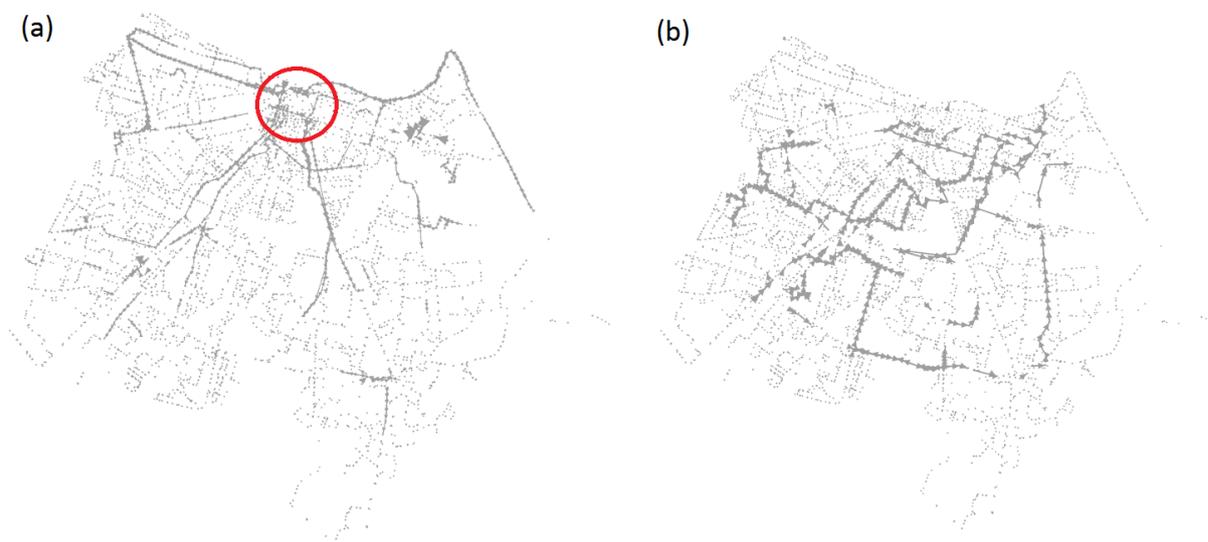

Figura 7. Trechos da rede onde foram encontrados gargalos e desperdício de recursos. Os trechos escuros destacam pontos de gargalo da rede em (a) e de desperdício de recursos em (b). Em (a) o círculo vermelho ainda destaca o centro comercial da cidade.

O índice, *IS*, proposto identificou que as linhas 503, 029, 014, 605, 725, 379, 087, 325, 316, 013, 130, 815, 030, 077, 011, 650, 755, 361, 612 e 075 têm trechos em seus itinerários onde os veículos se encontram lotados. Já *ID* identificou que as linhas 394, 315, 024, 220, 605, 709, 087, 013, 050, 311 e 317 estão em algum momento de seu itinerário praticamente vazias. Em validação *in loco* com profissionais da ETUFOR (Empresa de Transporte Urbano de Fortaleza) nos foi reportado que é conhecido que em alguns dias da semana as linhas 503, 029, 014, 605, 725, 379, 087, 325, 316, 013, 130, 815, 030, 077, 075 e 011 estão lotadas nos trechos destacados. Quanto as linhas 650, 755, 361 e 612 os profissionais se mostraram surpresos que as mesmas estejam lotadas, e informaram que as mesmas serão avaliadas em ações futuras. Sobre as linhas com poucos passageiros os profissionais preferiram não opinar. Foi justificado que é difícil saber se uma linha está com poucos passageiros em um determinado horário, pois esse fato não gera reclamação por parte dos usuários.

## 6. Considerações finais

Este trabalho explorou a oferta e a demanda do sistema de ônibus de Fortaleza, uma grande metrópole brasileira. Algoritmos e métricas de redes complexas foram usados para explorar o desequilíbrio entre o que o poder público oferece e o que a população necessita.

Em nível macro, as diferenças entre as redes de oferta e demanda em suas respectivas distribuições de densidade de pesos das arestas, distribuições cumuladas de peso das arestas e nível de modularidade, revelaram um desequilíbrio global no sistema. Em nível micro, foi desenvolvido um modelo que faz uso de técnicas de detecção de comunidades para identificar onde estão os gargalos e onde estão sendo desperdiçados recursos. Esse modelo se mostrou promissor mesmo em situações onde não é possível ter informações completas a respeito da demanda de passageiros.